\pdfoutput=1

\documentclass[11pt]{article}

\usepackage{xcolor}
\usepackage{times}
\usepackage{latexsym}
\usepackage{microtype}
\usepackage{graphicx}
\usepackage{subfigure}
\usepackage{booktabs} 
\usepackage{natbib}
\usepackage{graphicx}
\usepackage{amsmath}
\usepackage{amscd}
\usepackage{amssymb}
\usepackage{amsthm}
\usepackage{mathabx}
\usepackage{algorithmic}
\usepackage{algorithm}
\usepackage{booktabs}
\usepackage{multirow}

\usepackage{paralist}
\usepackage{color, colortbl}
\usepackage{enumitem}
\usepackage[normalem]{ulem}
\usepackage{stackengine}

\usepackage{naacl2021}

\usepackage{times}
\usepackage{latexsym}

\usepackage[T1]{fontenc}

\usepackage[utf8]{inputenc}

\usepackage{microtype}

\title{Towards Continual Learning for Multilingual Machine Translation via Vocabulary Substitution}

\author{Xavier Garcia \and Noah Constant \and Ankur P. Parikh \and Orhan Firat \\
         Google Research \\
         Mountain View \\
         California \\
         \texttt{{xgarcia,nconstant,aparikh,orhanf}@google.com}
         }

\begin{document}
\maketitle
\begin{abstract}
We propose a straightforward vocabulary adaptation scheme to extend the language capacity of multilingual machine translation models, paving the way towards efficient continual learning for multilingual machine translation. Our approach is suitable for large-scale datasets, applies to distant languages with unseen scripts, incurs only minor degradation on the translation performance for the original language pairs and provides competitive performance even in the case where we only possess monolingual data for the new languages.
\end{abstract}

\section{Introduction} \label{sec:intro}

The longstanding goal of multilingual machine translation \cite{firat16,johnson16,aharoni19,gu2018universal} has been to develop a universal translation model, capable of providing high-quality translations between any pair of languages. Due to limitations on the data available, however, current approaches rely on first selecting a set of languages for which we have data and training an initial translation model on this data jointly for all languages in a multi-task setup.  In an ideal setting, one would continually update the model once data for new language pairs arrives. This setting, dubbed in the literature as \emph{continual learning} \cite{ring1994continual,rebuffi2017icarl,kirkpatrick2017overcoming,lopez2017gradient}, introduces new challenges not found in the traditional multi-task setup, most famously \emph{catastrophic forgetting} \cite{mccloskey1989catastrophic}, in which the model may lose its previously-learned knowledge as it learns new language pairs. This situation is further complicated by the training procedures of standard tokenizers, such as Byte-Pair Encoding (BPE) \cite{sennrich2015neural} or Sentencepiece \cite{kudo18}, which necessitate access to monolingual data for all the languages considered before producing the vocabulary. Failing to comply with these requirements, one risks suboptimal segmentation rules which in the worst case could result in strings of entirely \texttt{<UNK>} tokens for text in a previously-unseen alphabet.

In this work, we investigate how vocabularies derived from BPE transform if they are rebuilt with the same settings but with additional data from a new language. We show in Section \ref{subsec:tok_overlap} that there is a large token overlap between the original and updated vocabularies. This large overlap allows us to retain the performance of a translation model after replacing its vocabulary with the updated vocabulary that additionally supports a new language.

Past works have explored adapting translation models to new languages, typically focusing on related languages which share similar scripts \cite{gu2018universal,neubig2018rapid,lakew2019adapting,chronopoulou2020reusing}. These works usually focus solely on learning the new language pair, with no consideration for catastrophic forgetting. Moreover, these works only examine the setting where the new language pair comes with parallel data, despite the reality that for a variety of low-resource languages, we may only possess high-quality monolingual data with no access to parallel data. Finally, unlike our approach, these approaches do not recover the vocabulary one would have built if one had access to the data for the new language from the very beginning.


Having alleviated the vocabulary issues, we study whether we are able to learn the new language pair rapidly and accurately, matching the performance of a model which had access to this data at the beginning of training. We propose a simple adaptation scheme that allows our translation model to attain competitive performance with strong bilingual and multilingual baselines in a small amount of additional gradient steps. Moreover, our model retains most of the translation quality on the original language pairs it was trained on, exhibiting no signs of catastrophic forgetting.

\section{Continual learning via vocabulary substitution} \label{subsec:spm_surgery}

\paragraph{Related works} Adapting translation models to new languages has been studied in the past.  \citet{neubig2018rapid} showed that a large multilingual translation model trained on a subset of languages of the TED dataset \cite{qi2018and} could perform translation on the remaining (related) languages. \citet{tang2020multilingual} was able to extend the multilingual translation model mBART \cite{liu2020multilingual} from 25 to 50 languages by exploiting the fact that mBART's vocabulary already supported those additional 25 languages. \cite{escolano2021bilingual} was able to add new languages to machine translation models by training language-specific encoders and decoders. Other works \cite{zoph16,lakew2018transfer,lakew2019adapting, escolano-etal-2019-bilingual} have studied repurposing translation models as initializations for bilingual models for a target low-resource language pair. Most recently \cite{chronopoulou2020reusing} examined reusing language models for high-resource languages as initializations for unsupervised translation models for a related low-resource language through the following recipe: build vocabulary $\mathcal{V}_{X}$ and a language model for high-resource language $X$; once data for low-resource language $Y$ arrives, build a joint vocabulary $\mathcal{V}_{X,Y}$ and let $\mathcal{V}_{Y|X}$ be the tokens from $Y$ that appear in $\mathcal{V}_{X,Y}$; substitute the vocabulary for the language model with the one given by $\mathcal{V}_{X} \cup \mathcal{V}_{Y|X}$ and use the language model as the initialization for the translation model.

\paragraph{Our approach}  
In this work, we are not only interested in the performance of our multilingual translation models on new language pairs, we also require that our models \emph{retain} the performance on the multiple language pairs that they were initially trained on. We will also be interested in how the performance of these models compares with those obtained in the oracle setup where we have all the data available from the start. The approaches discussed above generate vocabularies that are likely different (both in selection and number of tokens) from the vocabulary one would obtain if one had \emph{a priori} access to the missing data, due to the special attention given to the new language.  This architectural divergence will only grow as we continually add new languages, which inhibits the comparisons to the oracle setup. We eliminate this mismatch by first building a vocabulary $\mathcal{V}_{N}$ on the $N$ languages available, then once the new language arrives, build a new vocabulary $\mathcal{V}_{N+1}$ as we would have if we had possessed the data from the beginning and replace $\mathcal{V}_N$ with $\mathcal{V}_{N+1}$. We then reuse the embeddings for tokens in the intersection\footnote{Tokens shared between the two vocabularies are also forced to share the same indices. The remaining tokens are rewritten but we still reuse the outdated embeddings.} and continue training.

\begin{figure*}[h]
\includegraphics[width=\textwidth]{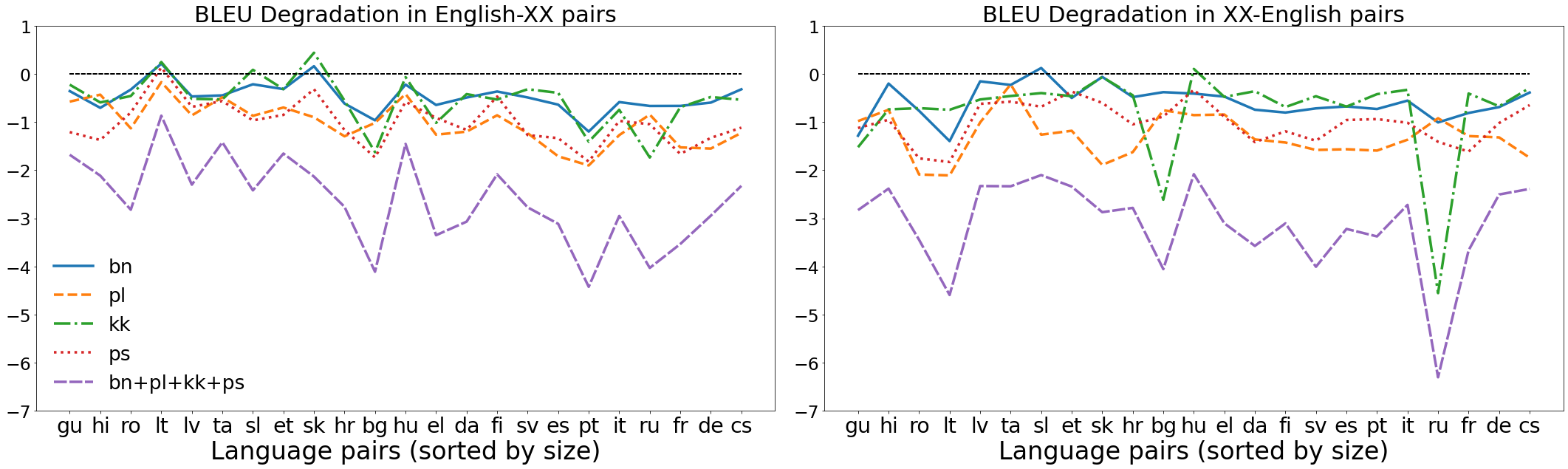}
\caption{\textbf{The degradation in BLEU from substituting vocabularies at inference}. The black dashed line represents the performance from the model trained with the modified vocabulary from the beginning, while the curves represent the BLEU scores from the original model using the new vocabulary at inference.}
\label{fig:adaptation_regression}
\end{figure*}

The success of our approach relies on the fact for large $N$ (i.e.~the multilingual setting), $\mathcal{V}_{N}$ and $\mathcal{V}_{N+1}$ are mostly equivalent, which allows the model to retain its performance after we substitute vocabularies. We verify this in the following section.
\section{Experiments}

In this section, we outline the set of experiments we conducted in this work. We first discuss the languages and data sources we use for our experiments. We then provide the training details for how we trained our initial translation models. 
Next, we  compute the token overlap between various vocabularies derived from BPE before and after we include data for a new language and empirically verify that this overlap is large if the vocabulary already suppots a large amount of languages. We then examine the amount of knowledge retained after vocabulary substitution by measuring the degradation of the translation performance on the original language pairs from replacing the original vocabulary with an updated one. Finally, we examine the speed and quality of the adaptation to new languages under various settings.

\paragraph{Languages considered} Our initial model will have to access to data coming from 24 languages\footnote{In alphabetical order: Bulgarian, Czech, Danish, German, Greek, English, Spanish, Estonian, Finnish, French, Gujarati, Hindi, Croatian, Hungarian, Italian, Lithuanian, Latvian, Portugese, Romanian, Russian, Slovak, Slovenian, Tamil.}. Our monolingual data comes primarily from the \emph{newscrawl} datasets\footnote{http://data.statmt.org/news-crawl/} and Wikipedia, while the parallel data comes WMT training sets and Paracrawl.  We will adapt our model to the following four languages: \textbf{Kazakh}, which is not related linguistically to any of the original 24 languages, but does share scripts with Russian and Bulgarian; \textbf{Bengali}, which is related to the other Indo-Aryan languages but possesses a distinct script; \textbf{Polish}, which is related to (and shares scripts with) many of the Slavic languages in our original set; \textbf{Pashto}, which is not closely related\footnote{Closest languages are in the Indic branch, but the Indic and Iranian branches split over 4000 years ago.} to any of the languages in our original set and has a distinct script. We provide an in-depth account of the data available for each language in the appendix.

\paragraph{Model configurations} We perform our experiments in JAX \cite{jax2018github}, using the neural network library FLAX\footnote{https://github.com/google/flax}. We use Transformers \cite{vaswani17} as the basis of our translation models. We use the Transformer Big configuration and a shared BPE model of 64k tokens with byte-level fallback using the Sentencepiece\footnote{We use 1.0 character coverage, split by whitespace, digits, and include a special token \texttt{MASK} for the MASS objective.} library. We used a maximum sequence length of 100, discarded all sequences longer than that during training. 

\paragraph{Sampling scheme} We train our models leveraging both monolingual and parallel datasets, following previous work \cite{siddhant-etal-2020-leveraging,garcia20}. We sample examples from monolingual and parallel sources with equal probability. Within each source, we use a temperature-based sampling scheme based on the numbers of samples of the relevant datasets with a temperature of 5 \cite{arivazhagan19a}.

\paragraph{Training objectives} We apply the MASS objective \cite{song19} on the monolingual data and cross-entropy on the parallel data. We used the Adam\cite{kingma14} optimizer, with an initial learning rate of 4e-4, coupled with a linear warmup followed by a linear decay to 0. The initial warmup took 1k steps, and the total training time was 500k steps. We also included weight decay with a hyperparameter of 0.2.  

\paragraph{Evaluation} We use beam search with a beam size of 4 and a length penalty of $\alpha = 0.6$ for decoding. We evaluate the quality of our models using BLEU scores \cite{papineni2002bleu}. We exclusively use detokenized BLEU, computed through sacreBLEU \cite{post-2018-call} for consistency with previous work and future reproducability.\footnote{BLEU + case.mixed + numrefs.1 + smooth.exp + tok.13a 

+ version.1.4.14}
\subsection{Transfer learning from vocabulary substitution}\label{subsec:tok_overlap}

\begin{table}
\centering
\small
\begin{tabular}{cccccc}
\toprule
\stackanchor{\# langs}{in base} & \emph{bn} & \emph{pl} & \emph{kk} & \emph{ps} & \stackanchor{\emph{bn}+\emph{pl}}{+\emph{kk}+\emph{ps}} \\ \midrule 
1 & 53.5\% & 47.0\% & 46.0\% & 47.8\% & 24.4\% \\
5 & 84.0\% & 80.8\% & 81.8\% & 80.2\% & 57.7\%  \\
10 & 90.3\% & 87.4\% & 89.3\% & 87.2\% & 70.9\%  \\
15 & 93.1\% & 91.8\% & 90.7\% & 90.5\% & 76.9\%  \\
20 & 94.8\% & 90.1\% & 93.0\% & 93.1\% & 79.2\%  \\
24 & 95.4\% & 94.3\% & 95.2\% & 93.5\% & 82.7\%  \\ \bottomrule
\end{tabular}
  \caption{\textbf{Percentage of token overlap between vocabularies before \& after the inclusion of a new language.} We denote the case where we add all the unseen languages by the column `bn+pl+kk+ps'.}
  \label{tab:tok_overlap}
\end{table}

\begin{table*}[!tb]
\small
\centering
\begin{tabular}{llcccccccc}
\toprule
 & \textbf{Model} & \multicolumn{2}{l}{\stackanchor{\, \emph{PMIndia}}{\, \emph{bn}$\leftrightarrow$\emph{en}}} &  \multicolumn{2}{l}{\stackanchor{\emph{newsdev2020}}{\emph{pl}$\leftrightarrow$\emph{en}}} & \multicolumn{2}{l}{\stackanchor{\emph{newstest2019}}{\emph{kk}$\leftrightarrow$\emph{en}}} & \multicolumn{2}{l}{\stackanchor{\emph{FLoRes devset}}{\emph{ps}$\leftrightarrow$\emph{en}}}   \\ \midrule
\multirow{3}{*}{\stackanchor{Original}{Vocabulary}}  & Unadapted & 0.0 & 0.0 & 2.4 & 4.0 & 0.7 & 2.2 & 0.0 & 0.0  \\ 
& xx monolingual \& parallel & 5.7 & 13.6 & 20.2 & 26.2 & 3.9 & 17.2 & 2.8 & 10.3 \\ 
& 4xx monolingual \& parallel & 5.3 & 15.1 & 18.3 & 25.0 & 2.7 & 15.8 & 2.3 & 8.4 \\\hline
\multirow{4}{*}{\stackanchor{Adapted}{Vocabulary}} & xx monolingual  & 0.0 & 1.7 & 13.9 & 24.3 & 0.9  & 19.0 & 0.0 & 6.5  \\
& xx monolingual (+BT)  & - & - & 21.3 & 24.1 & 4.7 & 19.5 & - & -    \\
&  xx monolingual \& parallel   & 10.0 & 27.2 & \textbf{21.5} & \textbf{27.5} & \textbf{5.9} & 20.2 & 6.6 & 15.1 \\
 &  4xx monolingual \& parallel   & \textbf{10.5} & 26.4 & 20.3 & 26.8 & 5.6 & \textbf{20.5} & \textbf{6.7} & \textbf{15.2}  \\ \hline
\multirow{2}{*}{\quad Oracle} 
& xx monolingual \& parallel  &  10.1 & \textbf{29.2} & 19.6 & 26.8 & 5.4 & \textbf{20.5} & 6.6 & 14.7 \\
& 4xx monolingual \& parallel &  10.0 & 28.6 & 18.9 & 26.4 & 5.4 & 20.3 & 6.2 & 14.4  \\ \bottomrule
\end{tabular}
\caption{\textbf{BLEU scores on the new languages.} The ``monolingual'' models have access to exclusively monolingual data for the new language(s), while ``monolingual \& parallel'' models add parallel data as well.  Models with ``xx'' add a single language, while ``4xx'' models add four languages together.}
\label{tab:adaptation_bleu}
\end{table*}

\begin{figure*}
\centering
\includegraphics[width=\textwidth]{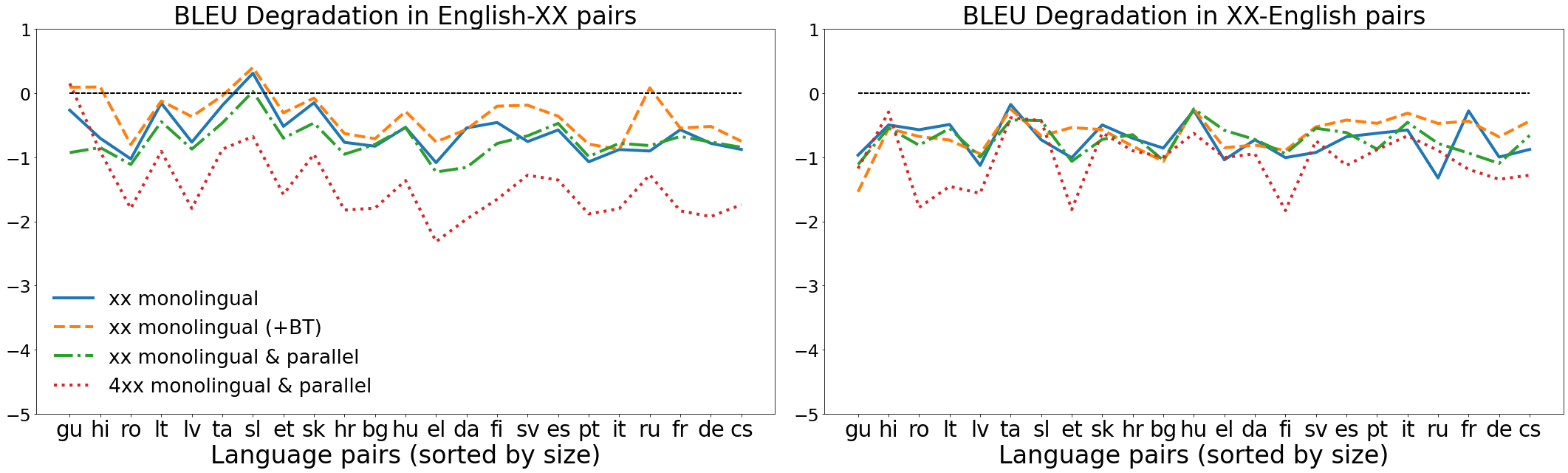}
\caption{\textbf{Measuring forgetting after adaptation.} The difference in BLEU for the original language pairs between the oracle model and models adapted to Kazakh.}
\label{fig:no_forgetting}
\end{figure*}

\paragraph{Measuring token overlap} We now examine the impact on the vocabulary derived from a BPE model upon the inclusion on a new language. We first build corpora consisting of text\footnote{We used 1 million lines of raw text per language.} from 1, 5, 10, 15, 20, and 24 of our original languages. For each corpus, we make copies and add additional data for either Bengali, Polish, Kazakh, Pashto, or their union, yielding a total of 30 corpora. We build BPE models using the same settings for each corpus, compute the token overlap between the vocabularies with and without the additional language, and report the results in Table \ref{tab:tok_overlap}. In the multilingual setting, we attain large token overlap, more than 90\%, even for languages with distinct scripts or when we add multiple languages at once. We extend this analysis to different vocabulary sizes and examine which tokens are ``lost'' in Appendix A.3.
\subsection{Evaluating translation quality and catastrophic forgetting} \label{subsec:adaptation}
\paragraph{Measuring the deterioration from swapping vocabularies at inference} 

To measure the amount of knowledge transferred through the vocabulary substitution, we compute the translation performance of our initial translation model with the adapted vocabularies \emph{without any additional updates}. For each new language, we compute the change in BLEU from the model with its original vocabulary and the one utilizing the adapted one and plot the results in Figure \ref{fig:adaptation_regression}. Notably, we only incur minor degradation in performance from the vocabulary substitution.

We now study the effect of introducing a new language into our translation model. We require an adaptation recipe which enjoys the following properties: \emph{fast}, in terms of number of additional gradient steps; \emph{performant}, in terms of BLEU scores on the new language pair; \emph{retentive}, in terms of minimal regression in the translation performance of the model on the original language pairs. 

Our solution: re-compute the probabilities for the temperature-based sampling scheme using the new data, upscale the probabilities of sampling new datasets by a factor then rescale the remaining probabilities so that their combined sum is one. We limit ourselves to either 15k or 30k additional steps (3\% and 6\% respectively of the training time for the original model) depending on the data available\footnote{We use 15k steps if we leverage both monolingual and parallel data for a single language pair. We use 30k steps if we only use monolingual data or if we are adapting to all four languages at once.} to ensure fast adaptation. We reset the Adam optimizer's stored accumulators, reset the learning rate to 5e-5 and keep it fixed. We provide more details in Appendix A.2. Aside from these modifications, we continue training with the same objectives as before unless noted otherwise. We include the results for oracle models trained in the same way as the original model but with access to both the adapted vocabulary \emph{and} the missing data. We compute the BLEU scores and report them in Table \ref{tab:adaptation_bleu}.

Our models adapted with parallel data are competitive with the oracle models, even when we add all four languages at once and despite the restrictions we imposed on our adaption scheme. For languages that share scripts with the original ones (Kazakh and Polish), we can also attain strong performance leveraging monolingual data alone, albeit we need to introduce back-translation \cite{sennrich2015improving} for optimal performance. We can also adapt the translation model using the original vocabulary, but the quality lags behind the models using the adapted vocabularies. This gap is larger for Bengali and Pashto, where the model is forced to rely on byte-level fallback, further reaffirming the value of using the adapted vocabularies.


To examine whether catastrophic forgetting has occured, we proceed as in Section \ref{subsec:tok_overlap} and examine the performance on the original language pairs after adaptation on the new data against the oracle model which had access to this data in the beginning of training. We present the results for the models adapted to Kazakh in Figure \ref{fig:no_forgetting}. All the models' performance on the original language pairs deviate only slightly from the oracle model, mitigating some of the degradation from the vocabulary substitution i.e.~compare the kk and bn+pl+kk+ps curves in Figure \ref{fig:adaptation_regression} to the curves in Figure \ref{fig:no_forgetting}.

Lastly, we compare our models with external baselines for Kazakh. We consider the multilingual model mBART \cite{liu2020multilingual} as well as all the WMT submissions that reported results on English $\leftrightarrow$ Kazakh. Of these baselines, only mBART and \cite{kocmi-etal-2018-cuni} use sacreBLEU which inhibits proper comparison with the rest of the models. We include them for completeness. We report the scores in Table \ref{tab:external_bleu}. Our adapted models are able to outperform mBART in both directions, and as well some of the weaker WMT submissions, despite those models specifically optimizing for that language pair and task.

\begin{table}[!tbp]
\small
\centering
\begin{tabular}{llcc}
\toprule
 & \textbf{Model} & \multicolumn{2}{l}{\stackanchor{\emph{newstest2019}}{\emph{kk}$\leftrightarrow$\emph{en}}}   \\ \midrule
\multirow{2}{*}{\stackanchor{Without}{$en\leftrightarrow kk$}} &  \emph{xx monolingual}  &  0.9  & 19.0 \\
& \emph{xx monolingual (+BT)} &  4.7 & 19.5    \\ \hline
\multirow{9}{*}{\stackanchor{With}{$en\leftrightarrow kk$}} & \citet{kocmi-bojar-2019-cuni} & 8.7 & 18.5 \\
& \citet{li2019niutrans} & 11.1 & \textbf{30.5} \\ 
& \citet{casas2019talp} & \textbf{15.5} & 21.0 \\
& \citet{dabre2019nict} & 6.4 & 26.4 \\
& \citet{briakou2019university} & - & 9.94 \\
& \citet{littell2019multi} & - & 25.0 \\
& mBART \cite{liu2020multilingual} & 2.5 & 7.4 \\
&  \emph{xx monolingual \& parallel}   &  5.9 & 20.2   \\
&  \emph{4xx monolingual \& parallel}   & 5.6 & 20.5  \\ \bottomrule
\end{tabular}
\caption{\textbf{BLEU scores on the new languages against external baselines.} The models in italics are ours.}
\label{tab:external_bleu}
\end{table}


\section{Conclusion}

We present an approach for adding new languages to multilingual translation models. Our approach allows for rapid adaptation to new languages with distinct scripts with only a minor degradation in performance on the original language pairs. 
\bibliography{anthology,custom}
\bibliographystyle{acl_natbib}

\appendix

\section{Appendix}
\label{sec:appendix}

\subsection{Data statistics and details}
We outline the counts, domains, test set, and BLEU scores of our original translation model on the 24 languages in Table \ref{tab:seen_langs}. We do the same for the unseen languages in Table \ref{tab:unseen_langs}. All the Paracrawl data is from v6.0.

\subsection{Adaption schemes}\label{app:finetune}

 We now explain in detail our configurations:

\paragraph{Monolingual data for a single language} In this case, we compute the probabilities following the temperature-based sampling scheme that we would have obtained had we computed with this data in the first place. Then we proceed to set the sampling probability of the new monolingual to 30\% and rescale the remaining probabilities so that they add up to 1. 

\paragraph{Monolingual data for a single language coupled with back-translation} In order to properly utilize back-translation, we first train the model for 10k step in the same fashion as the previous paragraph. Then, we use offline backtranslation on the new monolingual data to generate pseudo-parallel data. We then treat this data as authentic and include it in the model. We set the sampling probability of the pseudo-parallel data to be 10\%, we reset the sampling probability of the monolingual data to 10\%, and rescale the rest so that they sum up to 1. We then continue training for an additional 20k steps, amounting to a total of 30k steps. 
\paragraph{Monolingual \& parallel data for a single language} We multiply the probabilities of the new parallel data by a factor of 10, set the sampling probability of the monolingual data to 10\% then rescale the reamining probabilities so that they are normalized. We then train for 15k steps.

\paragraph{Monolingual \& parallel data for all four languages} We do not use the same scaling as before, since this would aggressively undersample the original language pairs. Instead, we first average the total probabilities for the new parallel data, multiply it by 5 and then assign this probability to each of the parallel datasets. We then fix the probability of sampling the new monolingual datasets to be 5\% each. We then train for 30k steps

\subsection{Token overlap analysis}\label{app:tok_overlap}

We first verify that the results in Table 1 apply for different vocabulary sizes. We compute analogous tables for vocabulary size of 32k and 128k tokens in Table \ref{tab:tok_overlap_32k} and \ref{tab:tok_overlap_128k} respectively.  

\begin{figure*}[!ht]
\centering
\includegraphics[width=0.5\textwidth]{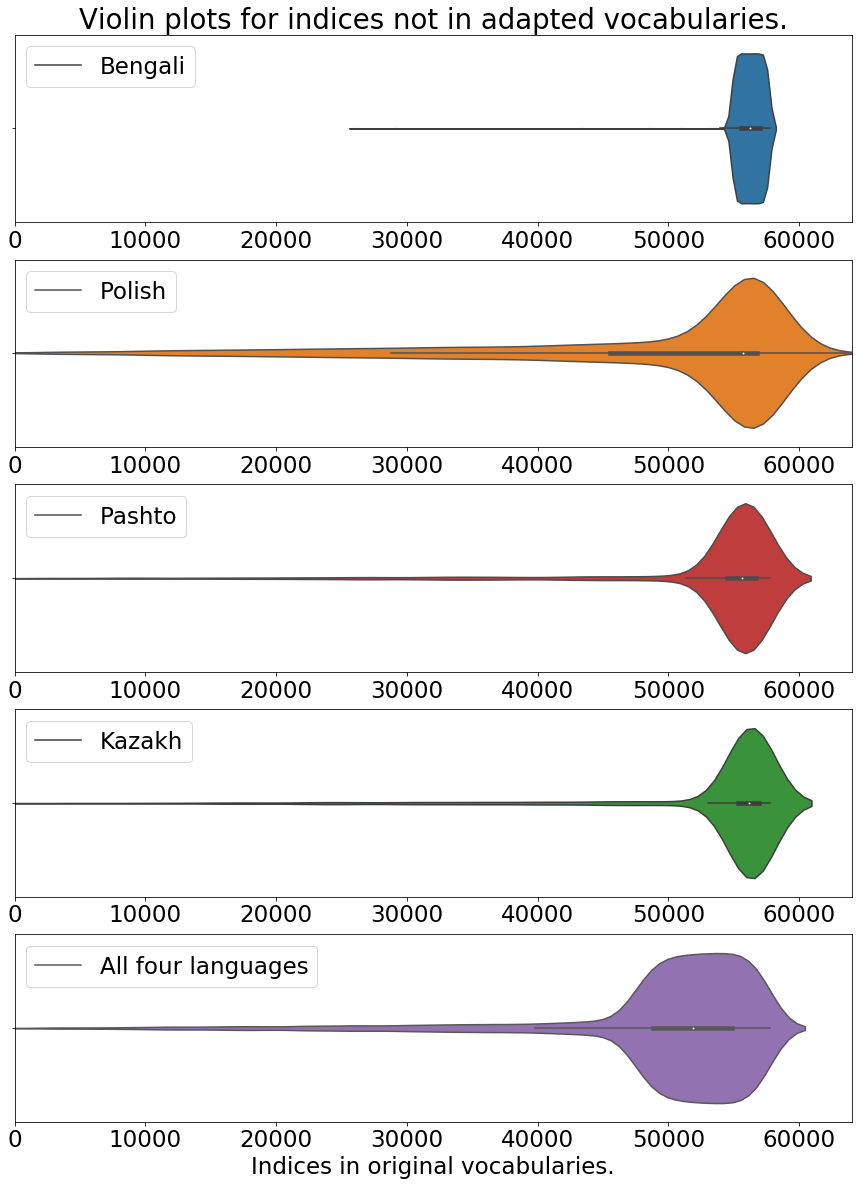}
\caption{\textbf{Violin plots for the indices in the original vocabulary that do not appear in the adapted vocabulary} Note that all language configurations, most of the indices that do not appear in the overlap are towards the infrequent side of the spectrum.}
\label{fig:quartile_ranges}
\end{figure*}

Next, we examine which tokens are lost during the vocabulary substitution. Since the Sentencepiece library does not provide an easy way to acquire frequency scores for BPE models after training, we instead use the order of the tokens as a proxy for the relative ranking obtained by sorting the tokens by frequency. For each language, we produce violin plots for the indices in the original vocabulary which are not in the adapted vocabulary for that language in Figure \ref{fig:quartile_ranges}. 

Critically, we observe that most of the tokens lost are towards the end of spectrum, suggesting that the model is mostly discarding infrequent tokens. Notably, it cannot discard the tail due to our requirement of full character coverage, which introduces a variety of rare Unicode characters as tokens that reside in the tail.

\begin{table}[!tbp]
\centering
\tiny
\begin{tabular}{cccccc}
\toprule
\stackanchor{\# languages}{in base model} & \emph{bn} & \emph{pl} & \emph{kk} & \emph{ps} & \emph{bn}+\emph{pl}+\emph{kk}+\emph{ps} \\ \midrule 
1 & 53.3\% & 49.3\% & 48.4\% & 47.7\% & 22.8\% \\
5 & 83.0\% & 81.1\% & 81.6\% & 78.0\% & 51.9\%  \\
10 & 89.7\% & 87.4\% & 88.9\% & 85.6\% & 65.1\% \\
15 & 92.8\% & 92.1\% & 90.2\% & 88.9\% & 72.1\%  \\
20 & 94.7\% & 90.3\% & 92.9\% & 92.2\% & 79.0\%  \\
24 & 95.6\% & 95.3\% & 95.5\% & 93.2\% & 83.8\%  \\ \bottomrule
\end{tabular}
  \caption{\textbf{Token overlap between vocabularies (consisting of 32k tokens) before \& after the inclusion of a new language.} }
  \label{tab:tok_overlap_32k}
\end{table}

\begin{table}[!tbp]
\centering
\tiny
\begin{tabular}{cccccc}
\toprule
\stackanchor{\# languages}{in base model} & \emph{bn} & \emph{pl} & \emph{kk} & \emph{ps} & \emph{bn}+\emph{pl}+\emph{kk}+\emph{ps} \\ \midrule 
1  & 53.5\% &  45.1\%   & 43.5\%   & 47.2\% &   21.1\% \\
5  & 85.1\% &  80.7\%  & 81.6\%   & 81.7\% &   54.0\%  \\
10  & 91.0\% &  87.5\%   & 89.3\%   & 88.4\% &   67.4\%  \\
15  & 93.6\% &  91.8\%   & 91.4\%   & 91.5\% &   74.6\%  \\
20  & 95.1\% &  90.3\%   & 93.3\%   & 93.5\% &   79.4\%  \\
24  & 95.5\% &  94.2\%   & 95.4\%   & 93.9\% &   82.8\%  \\ \bottomrule
\end{tabular}
  \caption{\textbf{Token overlap between vocabularies (consisting of 128k tokens) before \& after the inclusion of a new language.} }
  \label{tab:tok_overlap_128k}
\end{table}

\begin{table*}[!tbp]
\tiny
\centering
\begin{tabular}{ccccccccc}
\toprule
 Language & \stackanchor{Monolingual data}{(\# of lines)} & \stackanchor{Parallel data}{(\# of examples)} & \stackanchor{Domain}{(Monolingual data)} & \stackanchor{Domain}{(Parallel data)} & Test set & Language family & \stackanchor{BLEU}{en-xx} & \stackanchor{BLEU}{xx-en} \\ \midrule
Bg & 39610418 & 4111172 & NewsCrawl & Paracrawl & TED & Slavic & 32.43 & 35.77 \\
Cs & 81708712 & 64336053 & NewsCrawl & WMT & WMT'18 & Slavic & 18.42 & 28.60\\
Da & 4139992 & 6370432 & Wiki & Paracrawl & TED & Germanic & 38.81 & 42.87\\
De & 333313278 & 4508785 & NewsCrawl & WMT & WMT'14 & Germanic &23.63 & 30.38\\
El & 8332782 & 5298946 & NewsCrawl & Paracrawl & TED & Hellenic &29.03 & 34.40\\
Es & 53874815 & 15182374 & NewsCrawl & WMT & WMT'13 & Romance &31.74 & 33.23\\
Et & 5367030 & 2175873 & NewsCrawl & WMT & WMT'18 & Uralic &16.99 & 27.53\\
Fi & 21520558 & 6587448 & NewsCrawl & WMT & WMT'19 & Uralic &16.95 & 27.08\\
Fr & 87063385 & 40853298 & NewsCrawl & WMT & WMT'14 & Romance & 35.04 & 36.13\\
Gu & 816575 & 155798 & NewsCrawl & WMT & WMT'19 & Indo-Aryan &10.92 & 20.91\\
Hi & 23611899 & 313748 & NewsCrawl & WMT & WMT'14 & Indo-Aryan &13.36 & 18.98\\
Hr & 6814690 & 6814690 & NewsCrawl & Paracrawl & TED & Slavic &25.31 & 34.81\\
Hu & 40879784 & 4963481 & NewsCrawl & Paracrawl & TED & Uralic &15.90 & 24.25\\
It & 2836989 & 2747344 & NewsCrawl & Paracrawl & TED & Romance &31.87 & 36.59\\
Lt & 2836989 & 635146 & NewsCrawl & WMT & WMT'19 & Baltic &11.56 & 30.82\\
Lv & 11338472 & 637599 & NewsCrawl & WMT & WMT'17 & Baltic &17.16 & 22.69\\
Pt & 9392574 & 20677300 & NewsCrawl & Paracrawl & TED & Romance &33.25 & 41.79\\
Ro & 21033306 & 610320 & NewsCrawl & WMT & WMT'16 & Romance &27.18 & 36.92\\
Ru & 93827187 & 38492126 & NewsCrawl & WMT & WMT'19 & Slavic &22.20 & 34.70\\
Sk & 3040748 & 3303841 & Wiki & Paracrawl & TED & Slavic &22.59 & 29.52\\
Sl & 2669157 & 1923589 & Wiki & Paracrawl & TED & Slavic &21.06 & 25.73\\
Ta & 708500 & 736479 & NewsCrawl & WMT & WMT'20 & Dravidian & 6.29 & 12.06\\ \bottomrule
\end{tabular}
\caption{Details on the original 24 languages considered. For Tamil, we did not have access to the test set, so we only reported scores on the newstest2019, so we used newsdev2019 instead. The BLEU scores are from the our original translation model.}
\label{tab:seen_langs}
\end{table*}

\begin{table*}[!tbp]
\tiny
\centering
\begin{tabular}{ccccccc}
\toprule
 Language & \stackanchor{Monolingual data}{(\# of lines)} & \stackanchor{Parallel data}{(\# of examples)} & \stackanchor{Domain}{(Monolingual data)} & \stackanchor{Domain}{(Parallel data)} & Test set & Language family \\ \midrule
bn & 3918906 & 27584 & Newscrawl & PMIndia & PMIndia & Indo-Aryan \\
kk & 4032908 & 222424 & Newscrawl + Wiki Dumps & WMT & WMT & Turkic \\
pl & 3788276 & 5001447 & Newscrawl & WMT & WMT & Slavic \\
ps & 6969911 & 1134604 & Newscrawl + CommonCrawl & WMT & WMT & Indo-Iranian\\\bottomrule
\end{tabular}
\caption{Details on the additional 4 languages considered for adaptation. For Polish, we did not have access to the test set so we used the dev set instead.}
\label{tab:unseen_langs}
\end{table*}

\end{document}